\documentclass[%
 aip,
 amsmath,amssymb,
 reprint,%
]{revtex4-1}

\usepackage{graphicx}
\usepackage{dcolumn}
\usepackage{bm}

\usepackage[utf8]{inputenc}
\usepackage[T1]{fontenc}
\usepackage{mathptmx}
\usepackage{etoolbox}
\usepackage[colorlinks=true,citecolor=blue]{hyperref}

\usepackage[caption=false]{subfig}
\usepackage{xcolor}

\makeatletter
\def\@email#1#2{%
 \endgroup
 \patchcmd{\titleblock@produce}
  {\frontmatter@RRAPformat}
  {\frontmatter@RRAPformat{\produce@RRAP{*#1\href{mailto:#2}{#2}}}\frontmatter@RRAPformat}
  {}{}
}%
\makeatother
\begin{document}

\preprint{AIP/123-QED}

\title[Ordinal analysis of lexical patterns]{Ordinal analysis of lexical patterns}
\author{David S\'{a}nchez}%
\affiliation{Institute for Cross-Disciplinary Physics and Complex Systems IFISC (UIB-CSIC), E-07122 Palma de Mallorca, Spain}
\author{Luciano Zunino}%
\email{lucianoz@ciop.unlp.edu.ar}
\affiliation{Centro de Investigaciones Ópticas (CONICET La Plata-CIC-UNLP), 1897 Gonnet, La Plata, Argentina}
\affiliation{Departamento de Ciencias Básicas, Facultad de Ingeniería, Universidad Nacional de La Plata (UNLP), 1900 La Plata, Argentina}
\author{Juan De Gregorio}%
\affiliation{Institute for Cross-Disciplinary Physics and Complex Systems IFISC (UIB-CSIC), E-07122 Palma de Mallorca, Spain}
\author{Raúl Toral}%
\affiliation{Institute for Cross-Disciplinary Physics and Complex Systems IFISC (UIB-CSIC), E-07122 Palma de Mallorca, Spain}
\author{Claudio Mirasso}%
\affiliation{Institute for Cross-Disciplinary Physics and Complex Systems IFISC (UIB-CSIC), E-07122 Palma de Mallorca, Spain}
\date{\today}

\begin{abstract}
Words are fundamental linguistic units that connect thoughts and things through meaning. However, words do not appear independently in a text sequence. The existence of syntactic rules induces correlations among neighboring words. Using an ordinal pattern approach, we present an analysis of lexical statistical connections for eleven major languages. We find that the diverse manners that languages utilize to express word relations give rise to unique pattern structural distributions. Furthermore, fluctuations of these pattern distributions for a given language can allow us to determine both the historical period when the text was written and its author. Taken together, our results emphasize the relevance
of ordinal time series analysis in linguistic typology, historical linguistics and stylometry.
\end{abstract}

\maketitle

\begin{quotation}
Natural languages are systems where complex relations are established between a huge number of words. This leads to a plentiful variety of forms that substantialize linguistic rules. Surprisingly enough, we find that a handful of ordinal patterns suffices to reliably characterize any language. Moreover, statistical fluctuations of these patterns can shed light on both date and authorship identification. Our method utilizes the sequential nature of language, which enables to map a sufficiently long text into a time series of word rankings.
\end{quotation}

\section{\label{sec:intro}Introduction}

Despite its complexity, language seems to be organized with a few structural principles~\cite{hau02}.
For example, every language has a lexicon of thousands of words. These are basic elements with a particular meaning
which can be combined in utterances to transmit a full idea. Although the potential number of combinations can be overwhelmingly large, a statistical analysis of lexical frequencies shows a scaling behavior (Zipf's law~\cite{zipf}) that establishes an inverse proportion with respect to word rankings. This probability distribution holds for large corpora and many different languages~\cite{pia14} and has been linked to a cognitive principle of least effort in human communication~\cite{mandelbrot,fer03}.

Yet, the Zipf's law yields no information on the selection rules that govern grammatical arrangements
within a sentence. Indeed, words with the highest frequencies often operate with a purely syntactic purpose, such as
determiners (e.g., \textit{the} in English), prepositions (\textit{of}), conjunctions (\textit{and}) or pronouns (\textit{I}),
but unigram distributions like the Zipf's law cannot provide insight into the deep relationships formed
between function and content words to produce meaningful sentences. What is desirable, thus,
is to investigate distributions of bigrams, trigrams, etc.~\cite{ha09} to have a complete picture of the statistical
patterns that underlie human language.

At first sight the task looks formidable. If $N$ is the vocabulary cardinality, the number of distinct $n$-grams
is $N^n$. For a rough estimate of $N=10^4$, the possible combinations become exceedingly large already
for $n=3$ and cannot hence be statistically analyzed with the largest available resources
(e.g., the Google Books corpus~\cite{mic11} includes around $10^{11}$ tokens).
Even if one takes into account syntactic rules
that forbid certain combinations, the number would continue to be enormous. Here, we take an approach that
significantly simplifies the problem while revealing at the same time interesting linguistic patterns.

\begin{figure}[t]
\begin{center}
\includegraphics[width=0.49\textwidth, clip]{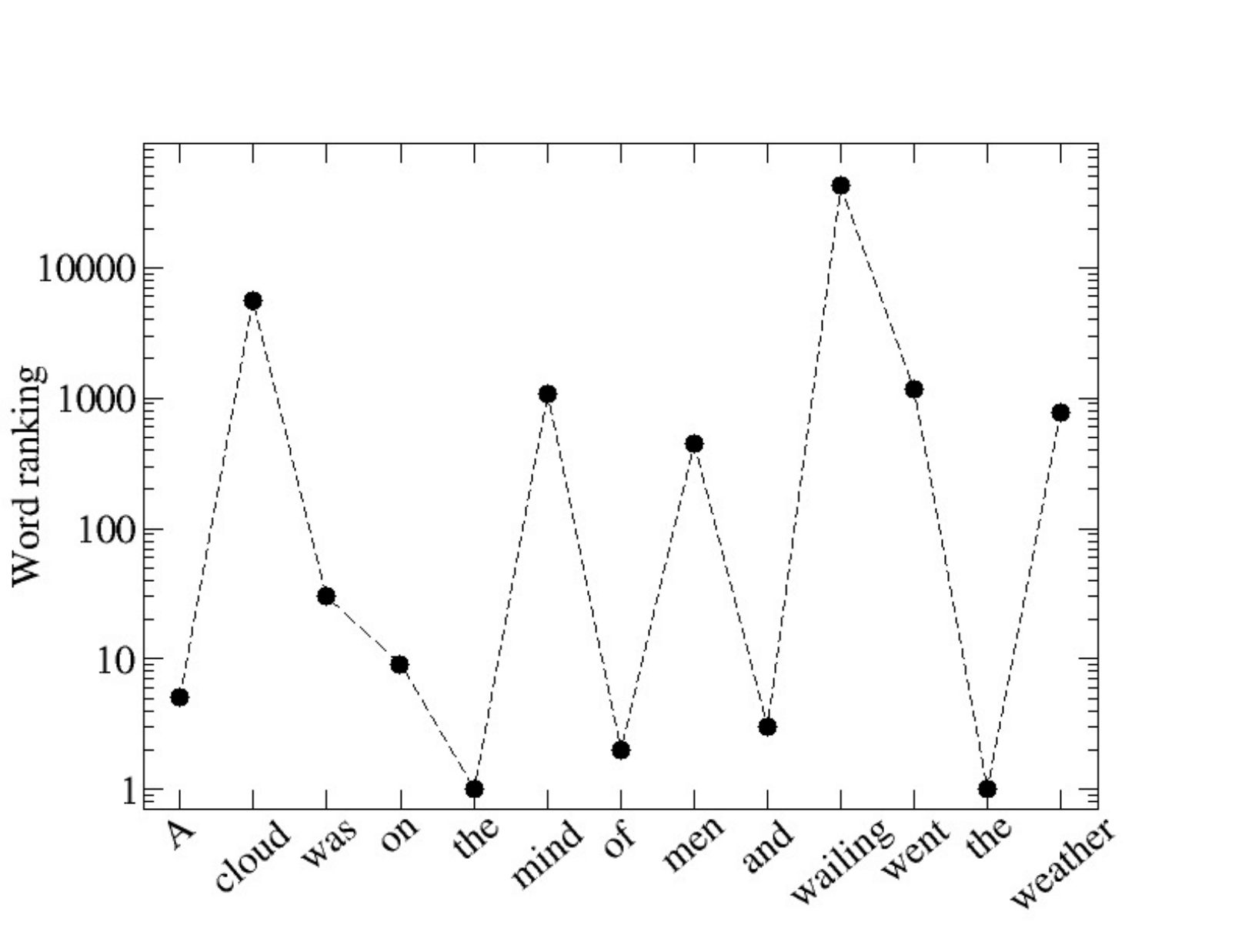}
\end{center}
\caption{{The text seen as a time series.} The frequency count for the tokens that appear in
\textit{The Man Who Was Thursday} establishes a frequency ranking for all word types therein. For illustrative purposes, we show the beginning of the novel.}
\label{fig_ches}
\end{figure}

Our approach is based on an ordinal analysis~\cite{bra02,zanin}. A text is viewed as a time series where the time dimension corresponds to the discrete position of the word inside the text. This perspective is accurate because language is sequential in nature: one word comes after the other. Let us consider the beginning of the \textit{The Man Who Was Thursday}, a 1908 novel by G. K. Chesterton: ``A cloud was on the mind of men and wailing went the weather\ldots". In Fig.~\ref{fig_ches} we plot the ranking of these words calculated from their absolute
frequencies within the novel as a function of position. It follows that \textit{the} is the top word type and appears
at the bottom of the time series while content words (\textit{cloud}, \textit{mind}, \textit{men}) possess
a much lower occurrence and come into the high part of the time series. As a consequence, any text portion in the book
consists of a succession of ups and downs as the story unfolds. Our aim is to study this ranking sequence rather than
the particular ranking value as the Zipf's law does. Below, we show that the distribution of
increasing and decreasing patterns contains valuable information not only about the language itself
but also about its history and the writer who generates the text. It is worth mentioning here that Sigaki~\textit{et al.}~\cite{sigaki2018} have recently shown that physics-inspired measures estimated from ordinal patterns distributions, when plotted in a complexity-entropy plane, are able to capture relevant information about paintings, their style and their temporal evolution. Moreover, these measures can be consistently connected with qualitative canonical concepts proposed by art historians to distinguish artworks. Furthermore, the same bidimensional representation space, but just using unigram word frequencies for estimating information theory quantifiers, has previously been applied with success to characterize plays and poems by Shakespeare and other English Renaissance authors\cite{rosso2009}. However, to the best of our knowledge, there has been no previous applications of ordinal analysis to texts. This is an interesting possibility that we explore in this work.

\section{\label{sec:method}Method}

Let $W$ be the number of words in a given text. We rank its words according to their absolute frequency
and convert the text into a sequence of rankings: $\mathcal{S}_r=\{r_1,r_2,\ldots,r_W\}$. This way, the $i$-th word in the sequence is replaced with its frequency ranking $r_i$. The rankings are calculated from each text separately. This guarantees that each word is assigned with a ranking.
Another possibility is to use a common ranking for all works under consideration (see Appendix~\ref{app_rank}), but our results are not significantly altered because $W$ is large for the texts considered in this work.
A word of caution is necessary for rare words~\cite{tan16} since it may be that two words with very low frequency share the same ranking. Whenever this happens we randomly modify the rankings of the affected words to make sure that in $\mathcal{S}_r$ two neighboring terms are never equal. In Appendix~\ref{app_seq} we give details of this procedure and prove that this modification does not affect the final results.

\begin{figure}[t]
\begin{center}
\includegraphics[width=0.49\textwidth, trim=5.cm 2.5cm 6.5cm 2.cm, clip]{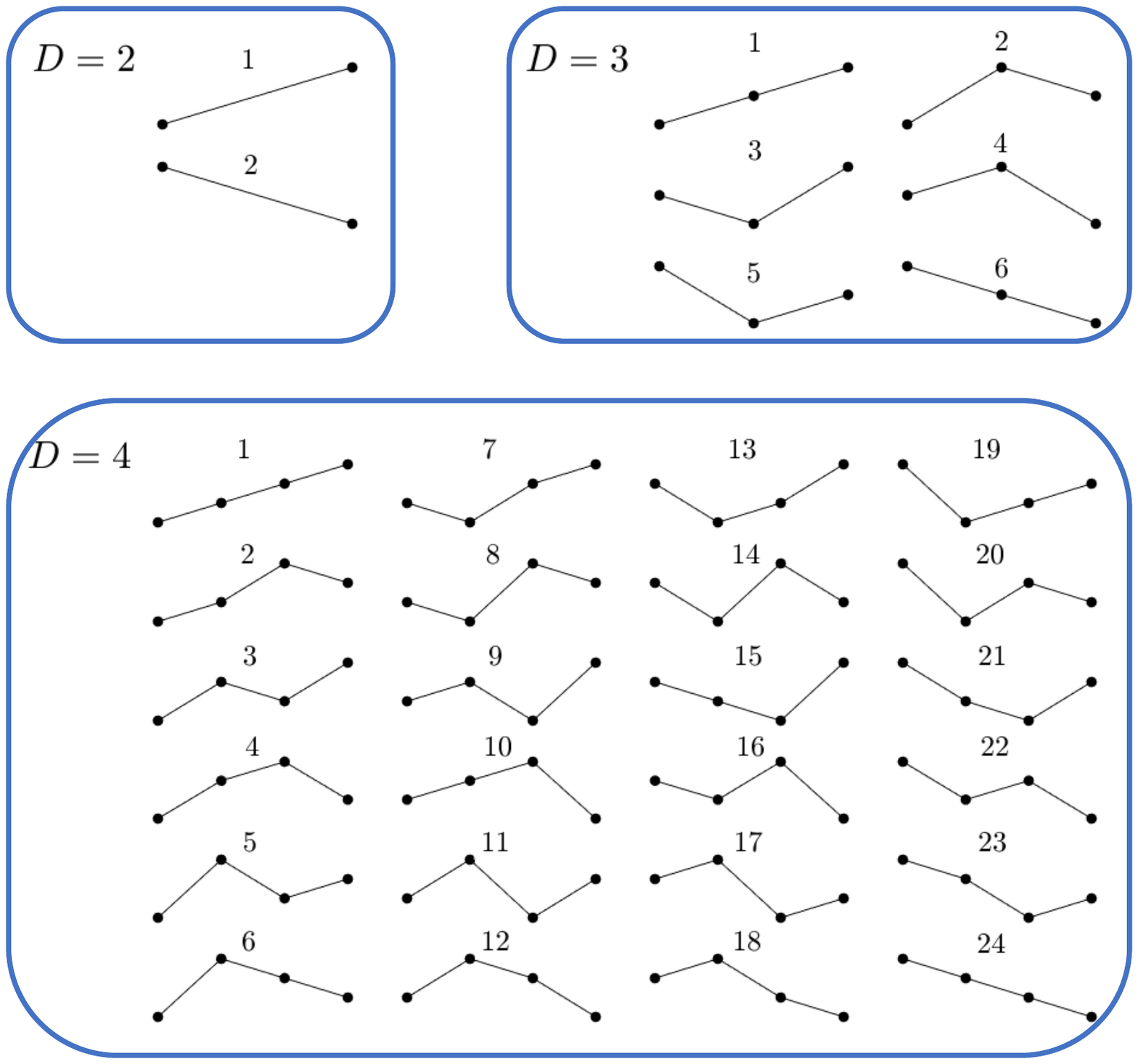}
\end{center}
\caption{{Ordinal patterns employed in this work.} We take embedding dimensions $D=2$ (top left), $D=3$ (top right) and $D=4$ (bottom). Each dot represents a value in a word ranking diagram such as the one in Fig.~\ref{fig_ches}.}
\label{fig_pat}
\end{figure}

Our objective is to obtain the ordinal pattern distribution for the text.
Depending on the embedding dimension
$D$ in the time series~\cite{tso92} there exist $D!$ ordinal patterns.
For instance, if $D=2$ as described above we have either an increasing
or a decreasing pattern between two consecutive words
with rankings $r_i<r_{i+1}$ in the first case and $r_{i+1}<r_i$ in the second case.
We plot a sketch of these in the top-left panel of Fig.~\ref{fig_pat}.
Then, for the data in Fig.~\ref{fig_ches} the ordinal pattern sequence becomes $\mathcal{S}_p=\{ 1, 2, 2, 2, 1, 2, 1, 2, 1, 2, 2,1\}$,
where we have assigned the symbols $1$ ($2$) to the increasing (decreasing) pattern.
For $D=3$ we have six possibilities, namely, $1$ ($r_i<r_{i+1}<r_{i+2})$,
$2$ ($r_i<r_{i+2}<r_{i+1}$), $3$ ($r_{i+1}<r_i<r_{i+2}$), $4$ ($r_{i+2}<r_{i}<r_{i+1}$),
$5$ ($r_{i+1}<r_{i+2}<r_{i}$) and $6$ ($r_{i+2}<r_{i+1}<r_i$),
see top-right panel in Fig.~\ref{fig_pat}. As a result,
the sequence in Fig.~\ref{fig_ches} can be symbolized as $\mathcal{S}_p=\{2,6,6,3,2,5,2,3,2,6,5\}$.
The procedure can be straightforwardly generalized to higher embedding dimensions (see bottom panel in Fig.~\ref{fig_pat} for the 24 ordinal patterns with $D=4$).

Additionally, one may consider the embedding delay $\tau \in \mathbb{N}$ that defines the time separation between the elements~\cite{tso92}. In the remainder of this paper, we take $\tau = 1$, which implies consecutive data, thus fulfilling the sequential property of language. As a consequence, the embedding dimension agrees with the number of items
in an $n$-gram. Another remark is in order. The pattern sequences $\mathcal{S}_p$
are generated allowing for overlaps between frequency rankings. Linguistically,
this implies that our method probes the phrase structure of the sentence.
This can be understood as follows. Quite generally, the pattern distributions
show text correlations between segments of length $(D-1)\tau$. We have checked
that for $\tau =3$ the results do not differ from a random sequence
obtained by shuffling the text words.
It follows that the method is sensitive to short-range correlations, unlike recent works that have paid attention to long-range correlations~\cite{sch93,ebe94,montemurro2002,alt12,alv06}. These short-range correlations occur at the phrase (syntagmatic) level~\cite{montemurro2002}. Below, we provide further evidence for this.

We perform an analysis in three different levels. In the macroscale, we contrast
the different ordinal pattern distributions across major language families~\cite{ethnologue}:
Indo-European (English, Spanish, French, German, Latin and Russian),
Afro-Asiatic (Somali), Niger-Congo (Xhosa), Turkic (Turkish), Austroasiatic (Vietnamese)
and Austronesian (Tagalog).
Our choice also allows for a broad variety of linguistic typologies. Since word order
plays a crucial role in our findings (see Sec.~\ref{sec:results} below),
we focus on the most common subject-verb-object (SVO) arrangements found in human language:
SVO (English, Spanish, French, Russian, Vietnamese, Xhosa), SOV (German, Latin, Somali, Turkish)
and VSO (Tagalog), which amount to 96\% of the existing typologies.
We only exclude East Asian families (Sino-Tibetan, Japonic, Koreanic)
because word boundaries are not clearly depicted in their written samples.
However, the number of selected languages suffices to support our findings.
To avoid possible differences due to the distinct nature of the analyzed texts
and allow for a fair comparison,
we need a single work, long enough, translated to the previously mentioned
languages. The Bible fulfills all these requirements, is publicly available
for natural language processing purposes~\cite{chr15}
and has already been employed in quantitative linguistics~\cite{meh17}.

In the mesoscale, we only consider one language (English) and examine its ordinal pattern distribution
over time. For definiteness, we bring our attention to the four periods into which
scholars traditionally divide the history of English: Old English, Middle English, Early Modern English
and Modern English~\cite{history}. We pick representative works for each period
(see Table~\ref{tab_list} in Appendix~\ref{app_lit}). 

Finally, the microscale is concerned with individual authors. We fix both the language
and the period (Modern English) and analyze a literary corpus~\cite{gutenberg} corresponding to
four writers: G.~K.~Chesterton, A.~C.~Doyle, H.~P.~Lovecraft and E.~A.~Poe.
Notice that the two most important varieties of English (British and American)
are equally represented with these authors. In Table~\ref{tab_list} of Appendix~\ref{app_lit}, we quote the five book titles for each of these writers employed for our microscale analysis.

All texts are tokenized using standard natural language processing toolkits~\cite{nltk}. This way, word forms are extracted and their rankings are straightforwardly calculated.
We neglect lemmatization because this affects a small amount of word types and because Zipf's law is preserved for lemmas~\cite{corral2015zipf}. Hence, we do not expect significant changes in the results.

\section{Results\label{sec:results}}

\subsection{Macroscale level\label{subsec:macro}}

\begin{figure}[t]
\begin{center}
\includegraphics[width=0.49\textwidth, trim=1.7cm 0cm 1.5cm 0.5cm, clip]{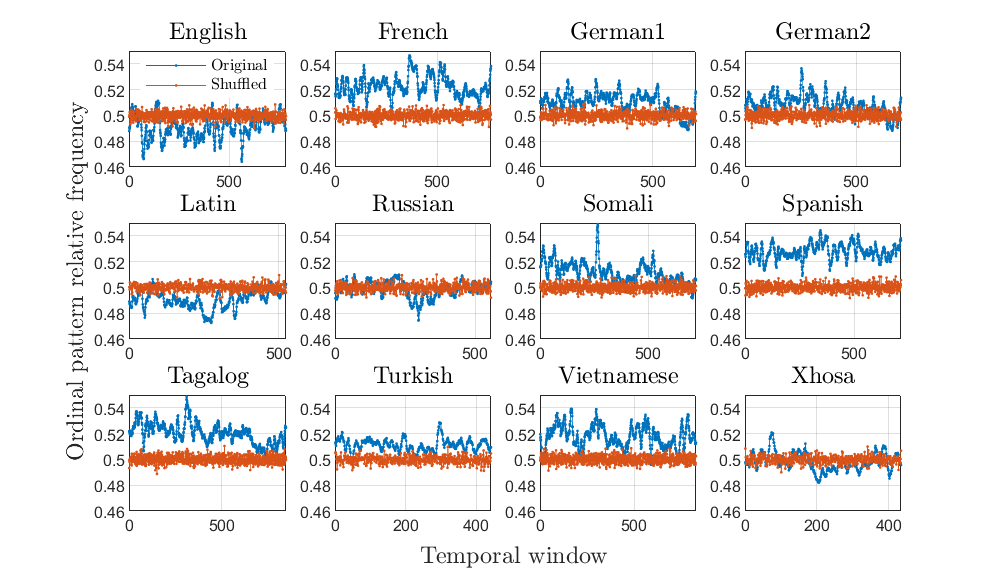}
\end{center}
\caption{{Relative frequency as a function of the text position
 for the Bible and the languages indicated above.} We take pattern~1 as defined in Fig.~\ref{fig_pat} for embedding dimension $D=2$. Each frequency is calculated for a temporal window of $10^4$ words. Then, the window is shifted $10^3$ time units until the text finishes. Labels in the horizontal axis indicate the different, consecutive, window series. Whereas the blue curves correspond to the original text, the red dots indicate a shuffled (random) realization generated by randomly varying the token positions.}
\label{fig_dyn}
\end{figure}

We start our analysis by showing with blue curves in Fig.~\ref{fig_dyn}
the normalized frequencies of the $D=2$ pattern 1 for the different Bibles.
(The frequency for the pattern 2 can be simply derived from probability normalization.)
Each frequency is calculated for a temporal window of $10^4$ words. Then, the window is shifted $10^3$
time units until the book finishes. For all languages, the signals appear stochastic
but clearly differ from a random sequence obtained by shuffling all the words (red curves).
In the latter case, the series also fluctuate but their mean is $0.5$ as should be.
In contrast, the expectation value for the original text is above or below $0.5$, depending
on the language, suggesting that the stationary probabilities contain correlations
entirely due to the word ordering dictated by the syntactic rules that operate in each human language.
Note that we depict two sequences for German, each
corresponding to a different Bible translation, showing that their dynamical behavior
do not show significant changes.

\begin{figure}[t]
\begin{center}
\includegraphics[width=0.49\textwidth, trim=1.7cm 0cm 1.5cm 0.5cm, clip]{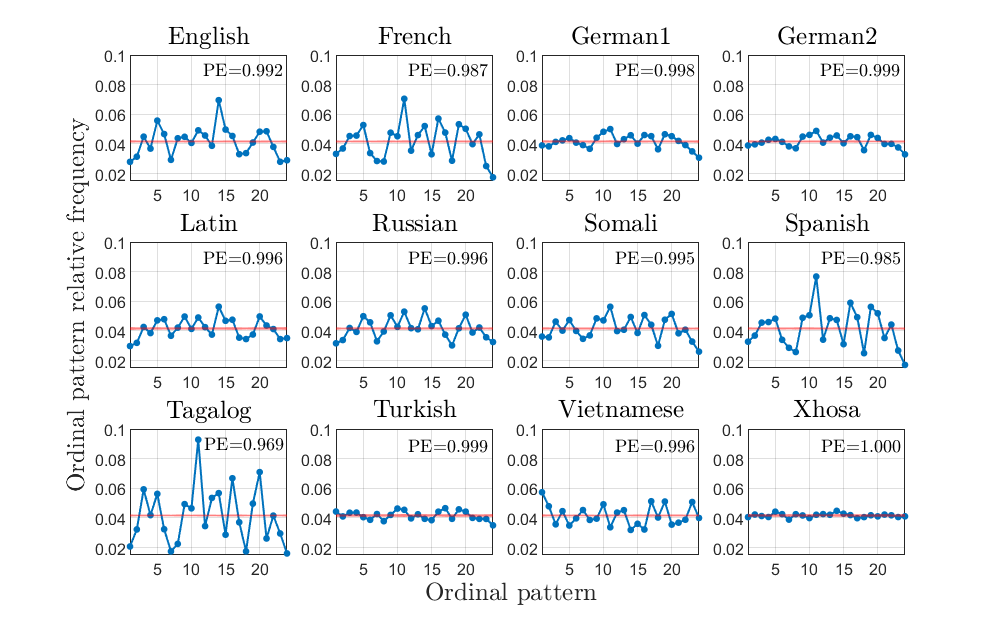}
\end{center}
 \caption{Macroscale results. We calculate pattern probability distributions at $D=4$ for the Bible in the indicated languages. In the $x$-axis we label the $D!=24$ possible patterns ordered as indicated in Fig.~\ref{fig_pat}. Similarly to Fig.~\ref{fig_dyn}, the blue dots correspond to the original texts while the red curves show results for an ensemble of 100 shuffled realizations. Normalized permutation entropy (PE) estimated values are also included in each panel.}
\label{fig_bib}
\end{figure}

We now compare in Fig.~\ref{fig_bib} (blue curves) the probability distributions for the observed stationary ordinal
patterns. Here, we choose a representative value for $D$
($D=4$ but the same conclusion is achieved for any other value
provided that $D! \ll W$).
Quite remarkably, we find that every language has its own fingerprint.
Admittedly, a few languages display similar histograms, such as French and Spanish
(both Romance languages), but this should not make us think that the distributions are determined
by the linguistic family. For instance, English and German are both Germanic languages
and show distinct probability functions. On the other hand, Xhosa (and possibly Turkish)
shows a uniform distribution close to the shuffled case, the latter shown as a red band with $3\sigma$ limits obtained after $100$ independent realizations.
This diversity of possible ordering of the time series under analysis can be quantified with the permutation entropy (PE in the insets of Fig.~\ref{fig_bib}), which is defined as the Shannon entropy of the ordinal pattern probability distribution. PE is the most representative and widely used ordinal descriptor. As seen, PE is essentially 1 for Xhosa and Turkish while the other languages show deviations form the uniform distribution.
We attribute this result to the fact that
Xhosa is a strongly agglutinative language where articles and prepositions are not typically
independent words but morphemes that join to root words. We further discuss this
particular effect in Appendix~\ref{app_seq}.
We also note that the results shown in Figs.~\ref{fig_bib}
are not altered if the text sentences are shuffled,
which is another proof that our ordinal approach only
detects short-range correlations that typically occur
among words inside a sentence
(see Appendix~\ref{app_shuf} for a more detailed discussion).
Finally, we do not observe any connection between the SVO order
and the pattern probabilities. The linguistic reason underlying the divergences
must be sought for somewhere else.

To gain further insight, we include in Table~\ref{tab_big} the most frequent bigrams and their $D=2$
associated pattern for both English (top) and Spanish (bottom). Whereas in English the pattern 2 is more common,
in Spanish the pattern with the highest probability is 1. What is the rationale for this difference?
If we examine the top bigrams we find that their parsing is preposition + determiner
(\textit{of the} in English or \textit{de la} in Spanish)
or determiner + noun (\textit{the Lord} in English or \textit{la tierra} in Spanish).
Therefore, their deep structures (in the generative grammar language sense~\cite{cry97}) do not differ.
It is instead the surface structure that determines the mean values for each pattern.
Since \textit{the} is the top word type in English, we find more instances of pattern 2
corresponding to the group preposition + determiner. Contrarily, this structure is built
in Spanish with the preposition \textit{de}, which is the top word type in this language
and, as a consequence, pattern 1 appears more often.
It is therefore not surprising that, as compared with Spanish, we derive an almost
equal distribution in French, which employ similar words for these functions
and with similar frequencies. Thus, the concrete pattern distributions
are not only caused by the syntactic rules but also by the diverse strategies that languages
employ to express these rules with the vocabulary at their disposal.
This does not preclude the existence of linguistic universals~\cite{gree63,mon11} but these are not captured
within our method. 

\begin{table}[t]
\caption{{ Bigrams with the largest frequencies.} We consider both the English (top) and Spanish (bottom) Bibles. We also include their $D=2$ ordinal pattern as labeled in Fig.~\ref{fig_pat}.}
\label{tab_big}
\subfloat{%
\begin{ruledtabular}
\begin{tabular}{lrc}
{ Bigram} & { Counts} & { Pattern} \\
\hline
 \textit{of the} & 11545 & 2 \\
 \textit{the lord} & 7036 & 1 \\
 \textit{and the} & 6278 & 2 \\
 \textit{in the} & 5031 & 2 \\
 \textit{and he} & 2794 & 1\\
 \hline
\end{tabular}
\end{ruledtabular}
}
\hfill
\subfloat{%
\begin{ruledtabular}
\begin{tabular}{lrc}
{ Bigram} & {Counts} & { Pattern} \\
\hline
 \textit{de la} & 4250 & 1 \\
 \textit{de los} & 3966 & 1 \\
 \textit{en el} & 2494 & 2 \\
 \textit{a los} & 2331 & 1\\
 \textit{la tierra} & 2202 & 1\\
 \hline
\end{tabular}
\end{ruledtabular}
}
\end{table}

\subsection{Mesoscale level\label{subsec:meso}}

\begin{figure}[t]

\subfloat{\includegraphics[width=0.24\textwidth, trim=0.cm 0cm 2.cm 0.cm, clip]{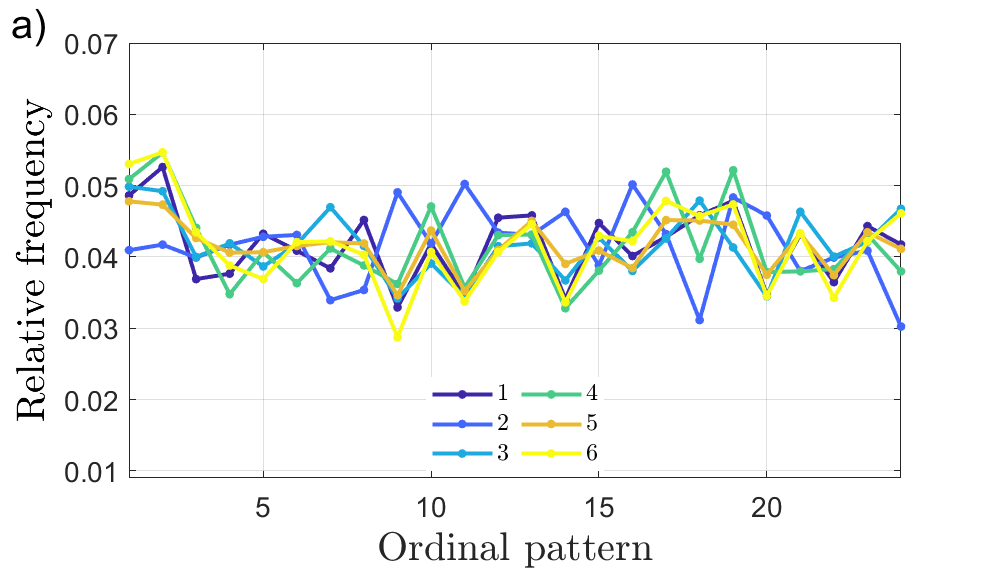}}
\hfill
\subfloat{\includegraphics[width=0.24\textwidth, trim=0.cm 0cm 2.cm 0.cm, clip]{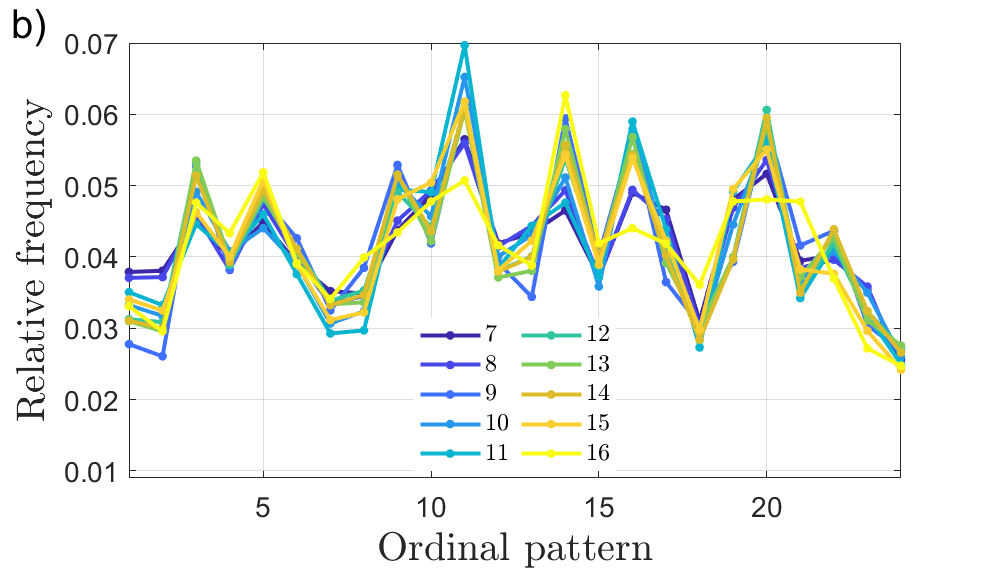}}
\newline
\subfloat{\includegraphics[width=0.24\textwidth, trim=0.cm 0cm 2.cm 0.cm, clip]{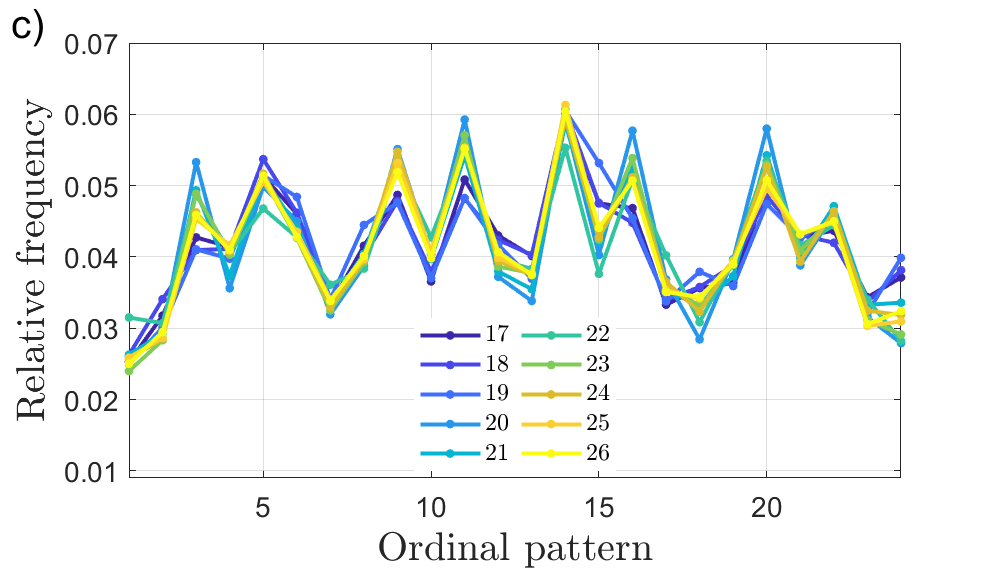}}
\hfill
\subfloat{\includegraphics[width=0.24\textwidth, trim=0.cm 0cm 2.cm 0.cm, clip]{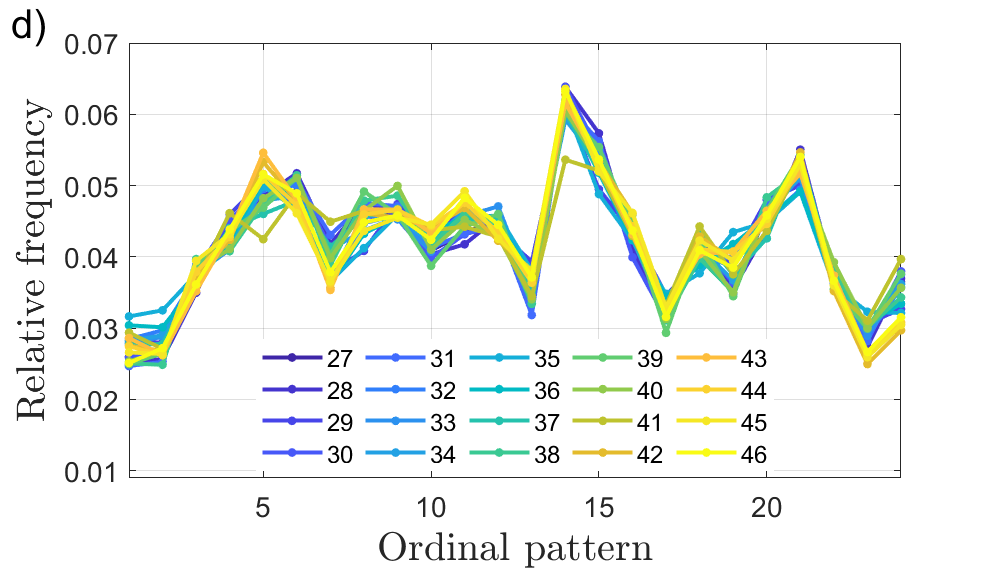}}

 \caption{Mesoscale results. We show the ordinal pattern probability distribution for embedding dimension $D=4$
 and different historical periods: a) Old English, b) Middle English, c) Early-Modern English and d) Modern English.
 Each curve corresponds to a single work or collection of works as listed in Table~\ref{tab_list} of Appendix~\ref{app_lit}. In the $x$-axis we label the $D!=24$ possible patterns ordered as indicated in Fig.~\ref{fig_pat}.}
\label{fig_his}
\end{figure}

Let us now discuss the mesoscale level.
It is well known that language changes with time. Then, we expect that pattern distributions will evolve along history. We illustrate this phenomenon in Fig.~\ref{fig_his}.
We take representative works for each historical period. In the Old English case [Fig.~\ref{fig_his}a)],
we plot the probability distribution function for the following works: \textit{Andreas} (curve 1), \textit{Anglo-Saxon Chronicle} (2),
\textit{Beowulf} (3), \textit{Christ} (4), \textit{Genesis} (5) and \textit{Guthlac} (6).
Despite the fact that these texts are quite short,
the distributions appear similar (with fluctuations due to the different lengths and genre).
For the Middle English [Fig.~\ref{fig_his}b)] we use \textit{Layamon's Brut} (to have texts of similar lengths we split this work in curves 7 and 8),
\textit{Canterbury Tales} (9, 10 and 11), \textit{Confession Amantis} (12, 13 and 14),
\textit{Book of the Knight of La Tour-Landry} (15) and \textit{Mandeville's Travels} (16).
It is interesting to see the evolution in the patterns from the Old [Fig.~\ref{fig_his}a)] to the Middle periods [Fig.~\ref{fig_his}b)]. The latter distributions appear more similar because of both the smaller time range of their period and a higher language standardization, a process that began in the Late Middle Ages.
The Early-Modern English corpus [Fig.~\ref{fig_his}c)] comprises works of Ben Jonson and those of Christopher Marlowe, Milton's \textit{Paradise Lost}, and Shakespeare's \textit{Tragedies} and \textit{Comedies}
(see Table~\ref{tab_list} in Appendix~\ref{app_lit} for their number identification). Finally, those chosen authors living in the Modern English period [Fig.~\ref{fig_his}d)]
were previously mentioned and their numbering are also included in Table~\ref{tab_list} in Appendix~\ref{app_lit}.
Clearly, there is an overall coherence among patterns belonging to the same time, which suggests that the traditional classification in periods has a lexical support.

\begin{figure}[t]
\begin{center}
\includegraphics[width=0.49\textwidth, trim=1.3cm 0cm 1.5cm 0.5cm,clip]{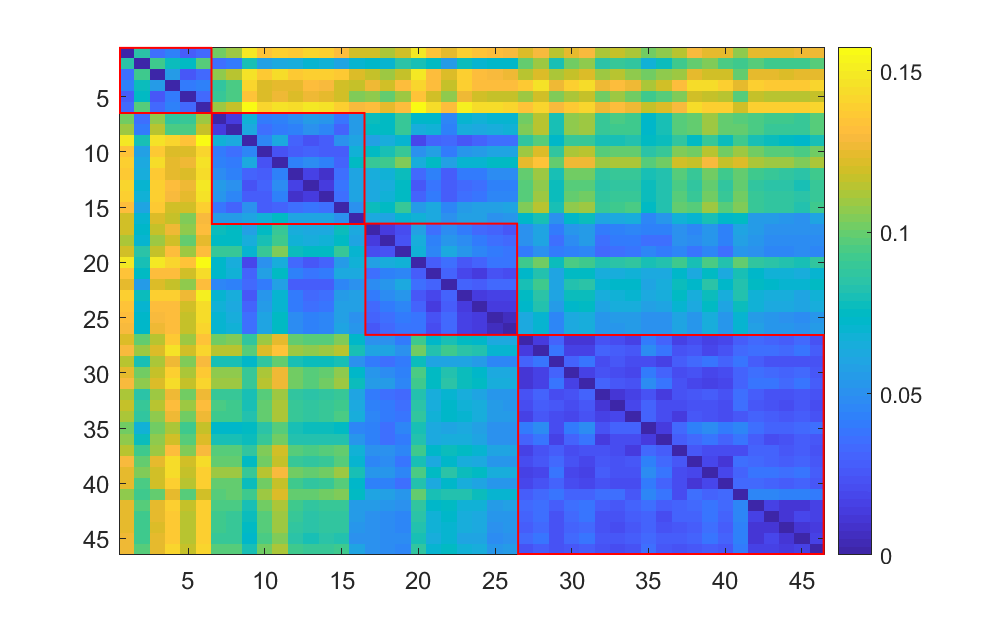}
\end{center}
 \caption{{ Permutation Jensen-Shannon distance for the mesoscale level.} The distance is determined between pairwise probability distributions shown in Fig.~\ref{fig_his} for $D=4$. Darker (lighter) colors indicate a small (large) distance, which quantifies the lexical (in the ordinal analysis language) difference among literary works belonging to distinct English historical periods. Red lines are included to help guide the eye.}
\label{fig_his2}
\end{figure}

This is better seen when one calculates the permutation Jensen-Shannon
distance~\cite{zunino2022} between distributions and plots this distance for $D=4$ as in Fig.~\ref{fig_his2}. As compared with the permutation entropy employed earlier in this work, the permutation Jensen-Shannon distance is more efficient to detect small differences between probability distributions. Remarkably,
we observe four dark areas that correspond to the four historical periods, indicated with red lines.
The distinction is clear between Old, Modern and the cluster formed by
Middle English and Early-Modern English, between which the transition
is less clear. This is because the Early-Modern English spans a period
between the Renaissance, when the medieval forms were still popular,
and the 17th century, when English conventions were approaching those of the
Modern period. Interestingly, there exist individual deviations from the historical
pattern. For instance, \textit{Anglo-Saxon Chronicle} (work 2) appears to be close to the Middle English
cluster whereas \textit{Paradise Lost} (work 22) would be more suitable
to be classified in the previous stage (Middle English), probably due to Milton's intentionally
archaic style. Another exception is Jonson (works 17, 18 and 19), whose style is better categorized
within the Modern period. We highlight that despite the method's simplicity
we are not only able to correctly place literary works in their composition period but also detect singular departures assignable to particular style features.

\subsection{Microscale level\label{subsec:micro}}

\begin{figure}[t]
\begin{center}
\includegraphics[width=0.49\textwidth, trim=1.3cm 0cm 1.5cm 0.5cm,clip]{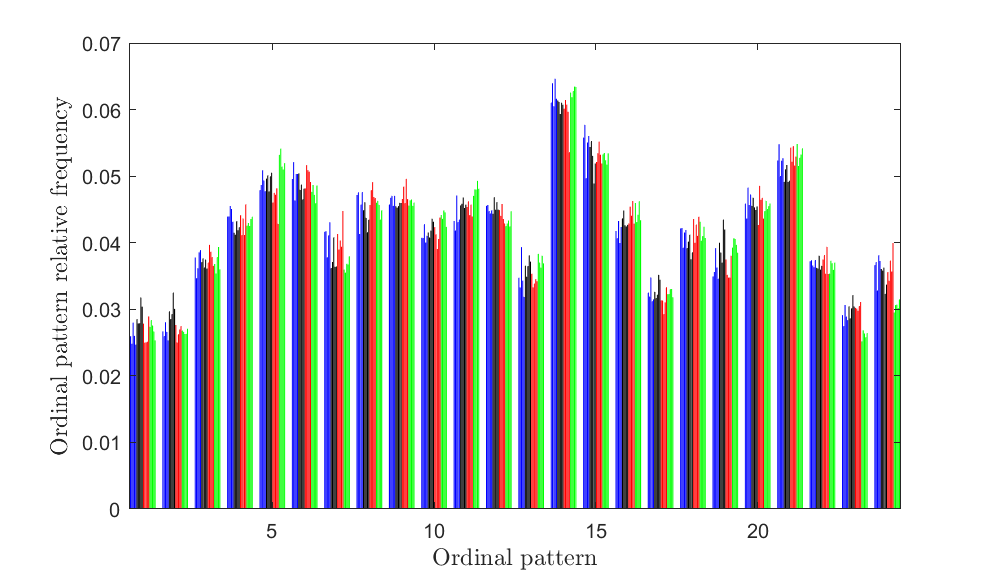}
\end{center}
 \caption{{Microscale results.} Histograms showing $D=4$ pattern probability distributions for four different authors from the same historical period (Modern English):
 G. K. Chesterton (blue), A. C. Doyle (black), H. P. Lovecraft (red) and E. A. Poe (green). On average the distributions are similar with fluctuations caused by author stylistic differences.}
\label{fig_fluc}
\end{figure}

\begin{figure}[t]
\begin{center}
\includegraphics[width=0.49\textwidth, trim=1.3cm 0cm 1.5cm 0.5cm,clip]{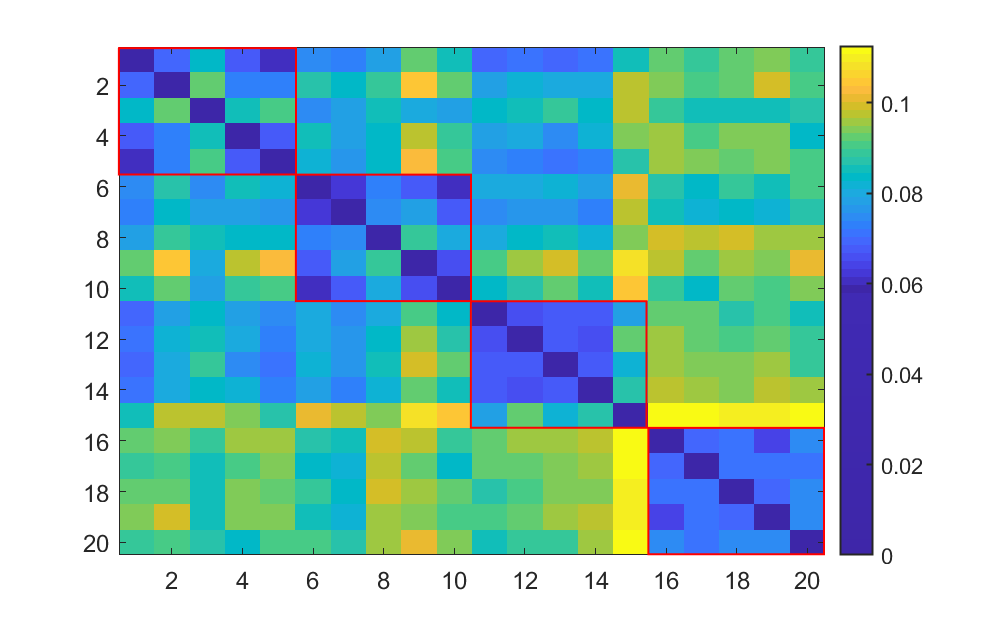}
\end{center}
 \caption{{Permutation Jensen-Shannon distance for the microscale level.} We determine the distance between the $D=6$ pattern probability distributions for the works of four English Modern authors (Chesterton, Conan Doyle, Lovecraft and Poe). The axes correspond to the numerical identification in parentheses of Table~\ref{tab_list} of Appendix~\ref{app_lit}. Red lines are included to help guide the eye.}
\label{fig_matrix}
\end{figure}

Previous results are encouraging because they show that on top of a common background
which characterizes written works in a given language there may exist fluctuations large enough
that allow us to determine the author of a set of texts. In fact, a few subclusters with smaller distances can be distinguished in Fig.~\ref{fig_his2} for works of the same writers. This is particularly evident for Layamon's, Gower's, Jonson's and Shakespeare's works. We now pursue this idea by further analyzing the last historical period (microscale). In Fig.~\ref{fig_fluc} we depict the ordinal
patterns and their probability frequencies for $D=4$. Fluctuations are seen
in the slight differences within the histogram. Then, we assess the pairwise distribution distance and plot the resulting matrix in Fig.~\ref{fig_matrix}.
To obtain a more efficient discrimination between the texts, thus amplifying
the fluctuations shown in Fig.~\ref{fig_fluc}, we set the embedding dimension to $D=6$, which allows for 720~patterns.
Strikingly, we observe in Fig.~\ref{fig_matrix} that each writer forms a cluster of his own, indicated with red lines. The largest distance takes place between
Poe (works numbered between 16 and 20) and the rest, perhaps due to Poe's highly mixed style.
Here, we add an important caveat: the microscale is the most sensitive situation and
Fig.~\ref{fig_matrix} is only a proof of concept. To use this technique in author
attribution tasks~\cite{gri07} would require better refined analyses that fall beyond the scope of the present work.

\section{Conclusion\label{sec:conclusion}}

To sum up, we have shown that the analysis of ordinal patterns is a powerful method that permits to distinguish
(i) language, (ii) historical period and (iii) single authors. 
First, every language has a characteristic fingerprint in terms of a statistical
distribution for symbolic patterns. The observed patterns emerge from a combination of the syntactic rules
that shape each language and the way that this language articulates those rules.
Second, a careful view of the pattern distribution provides useful information on the historical
period when the text was produced. Third,
since patterns have their origin in the greatly diverse
procedures with which human languages embody the syntactic relations that constrain word combinations, the distributions show fluctuations that can be traced back to texts written by single authors.

The procedure discussed here has obvious limitations, the most important of which concerns semantics.
Since every word is replaced with its ranking value in a table of frequencies, the symbolic patterns are agnostic
with regard to meaning. However, this is the same limitation that takes place in all information-theoretic
approaches to language since Shannon's theory of communications~\cite{shannon}. 

Our findings bode well for possible applications
of our method. We envisage implementations in stylometry studies that seek a correct authorship attribution
or in forensic linguistics for legal cases where linguistic data play a decisive role. Another interesting
application would aim at the detection of speech impairments in individuals. Additionally, our idea could be useful within the sociolinguistics realm
(characterization of dialects, registers and idiolects). An interesting avenue of future research would be to apply our method to spoken corpora and investigate whether there exist differences with the text corpora employed in this work.

\begin{acknowledgments}
We thank M. Zanin for useful comments.
Financial support has been received from MCIN/AEI/10.13039/501100011033, the Fondo Europeo de Desarrollo Regional (FEDER, UE) through project APASOS (PID2021-122256NB-C21) and the Program for Centers and Units of Excellence in R\&D, María de Maeztu project CEX2021- 001164-M/10.13039/501100011033, and the CAIB through the ITS2017-006 project (No. PDR2020/51). L.Z. acknowledges the financial support from Consejo Nacional de Investigaciones Científicas y Técnicas (CONICET), Argentina.
\end{acknowledgments}

\section*{Data Availability Statement}

The data that support the findings of this study are openly available in Figshare at http://doi.org/10.6084/m9.figshare.21762947.v1.

\appendix

\section{Common ranking}\label{app_rank}
In this appendix, we calculate the ranking sequences differently.
We consider a large corpus and arrange its words based on their
occurrences. The corresponding rankings are then used
to determine the ordinal patterns. The advantage of this approach
is that all literary works are symbolized using the same ranking.
The limitation is that word types that do not appear in the corpus
cannot be assigned to a definite ranking and therefore
not all patterns consist of consecutive words. However,
we do not see a significant difference between both methods.

\begin{figure}[t]

\subfloat{\includegraphics[width=0.24\textwidth, trim=0.cm 0cm 2.cm 0.cm, clip]{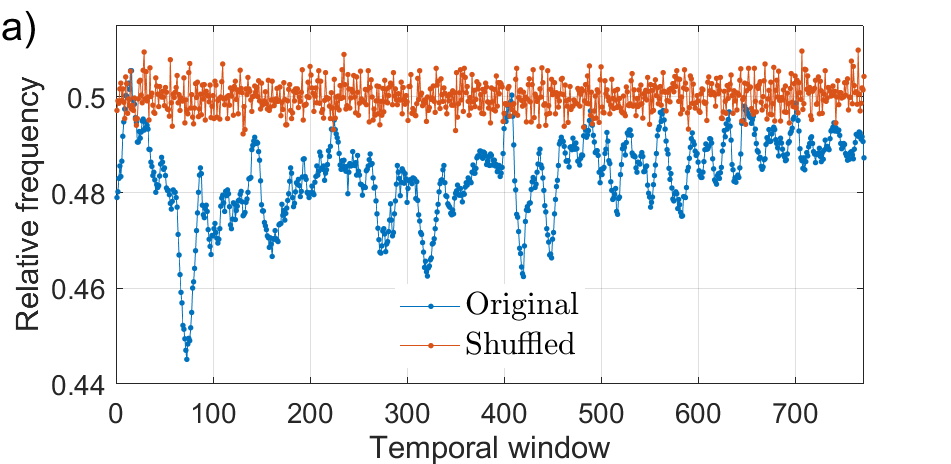}}
\hfill
\subfloat{\includegraphics[width=0.24\textwidth, trim=0.cm 0cm 2.cm 0.cm, clip]{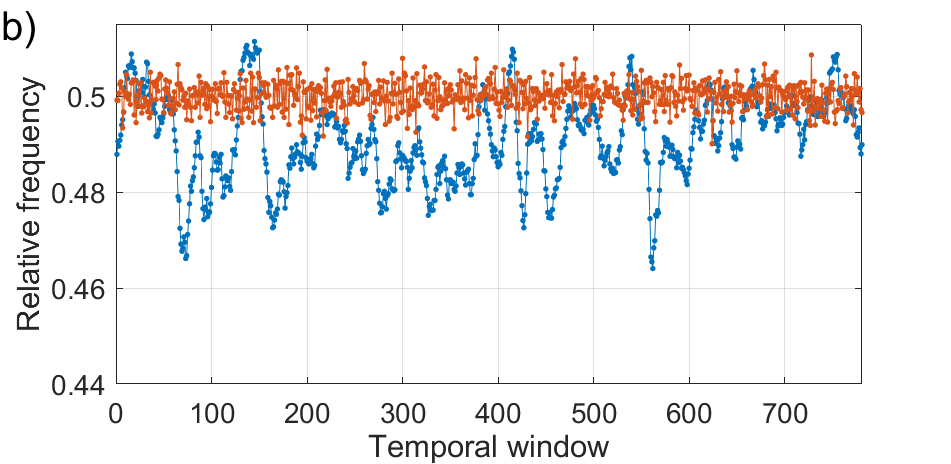}}
\newline
\subfloat{\includegraphics[width=0.24\textwidth, trim=0.cm 0cm 2.cm 0.cm, clip]{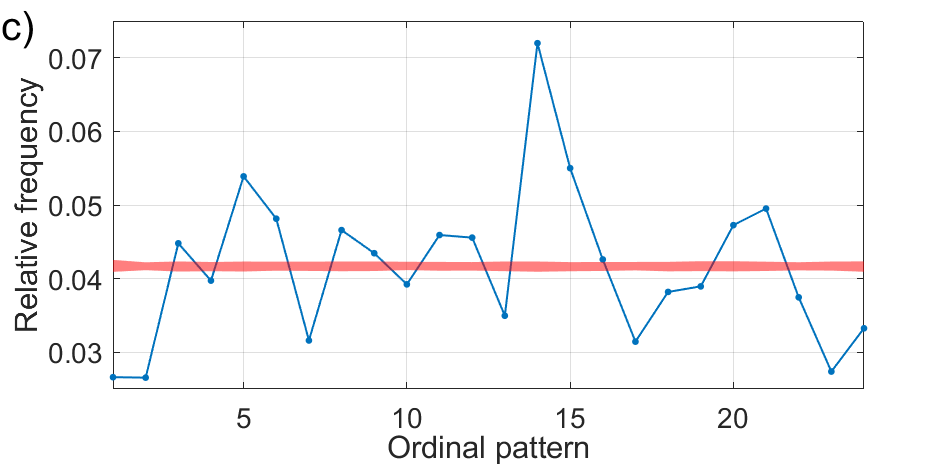}}
\hfill
\subfloat{\includegraphics[width=0.24\textwidth, trim=0.cm 0cm 2.cm 0.cm, clip]{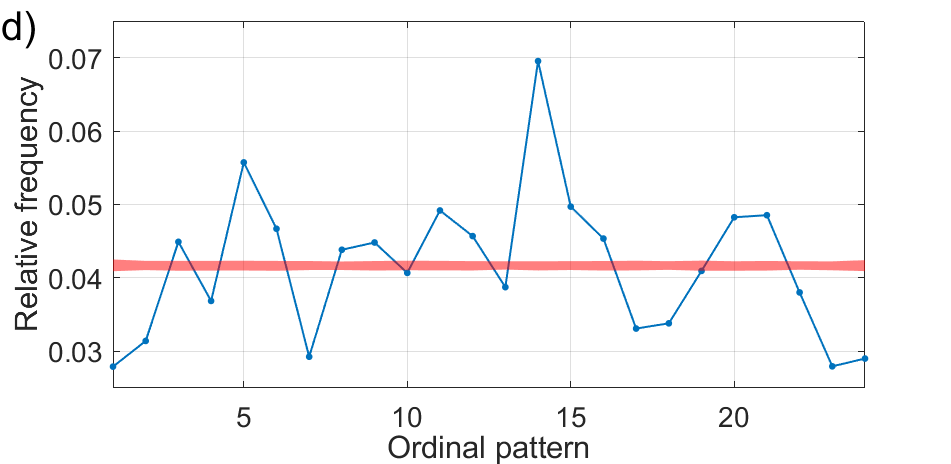}}

\caption{Dynamical behavior of the $D=2$ pattern 1 (blue curves) for the English Bible when the ranking sequences are generated by using a) a common and b) its own corpus. The corresponding ordinal pattern probability distributions for $D=4$ are displayed in c) and d), respectively. Results obtained when words are shuffled (red curves) are also included only for reference purposes. As in Sec.~\ref{subsec:macro}, in the dynamical panels we only include a single shuffling realization to avoid finite size effects but in the distribution panels the red curves are indeed bands with $3\sigma$ limits calculated after 100 realizations.}
\label{fig_GC}
\end{figure}

It suffices to illustrate this fact with a single language (e.g., English). We have checked
that our conclusions are unaltered for different languages.
The English word frequency list contains the 1/3 million most frequent
words~\cite{nor09} built from the Google Books Corpus (GBC)~\cite{google}.
In Fig.~\ref{fig_GC}a) we depict the symbol dynamics for $D=2$
obtained from the GBC ranking, comparing with the original series,
which is reproduced in Fig.~\ref{fig_GC}b) from the top left panel in Fig.~3. We find that the dynamical patterns resemble each other
although the peak amplitudes differ. This is expected because the strength of the fluctuations depends on the word frequencies, which in turn
are calculated from different corpora. However, the probability
distributions are almost unaltered. We show this in
Fig.~\ref{fig_GC}c) for the GBC side by side with
Fig.~\ref{fig_GC}d), which is replicated from the top left panel in Fig.~4. This demonstrates the robustness of our method
for alternative corpora provided that the size of the corpus is sufficiently large.

\section{Sequences with equal rankings} \label{app_seq}
\begin{figure}[t]
\begin{center}
\includegraphics[width=0.49\textwidth, clip]{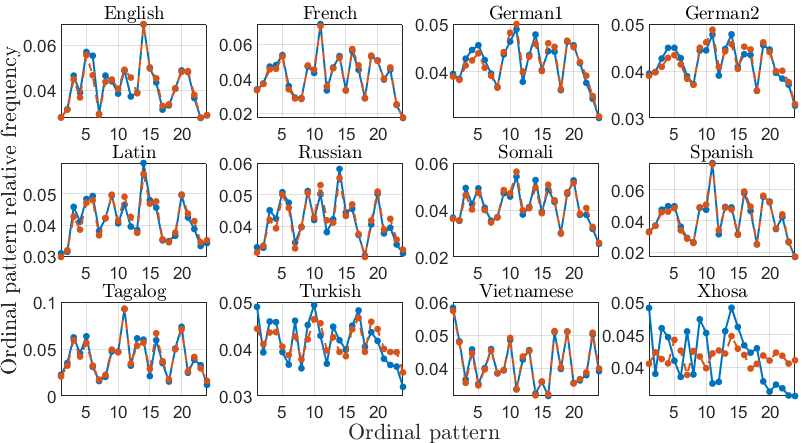}
\end{center}
 \caption{Distribution of linguistic ordinal patterns for embedding dimension $D=4$ for the original text sequences (blue lines) and when a small random number is added to break ranking equalities (dashed red lines).}
\label{fig_noise}
\end{figure}

If a sequence of $k$ words have the same ranking $r_{i}<r_{i+1}=r_{i+2}=\dots=r_{i+k}<r_{i+k+1}$, then we modify randomly those rankings by adding to each of $r_{i+1},\dots,r_{i+k}$ a uniform random number in the interval $(-(r_{i+1}-r_i),r_{i+k+1}-r_{i+k})$.
Here, we provide evidence that our procedure of breaking ranking
ties does not affect the main results except for a case that deserves
attention. In Fig.~\ref{fig_noise} we plot with solid blue lines
the pattern distributions when two
words are allowed to have the same ranking.
If this happens, equal rankings are sorted according to their temporal order. In addition, in Fig.~\ref{fig_noise} we reproduce
with red dashed lines the case as in Fig.~4.
Clearly, adding a small random number preserves the general structure of the
probability distributions. The case
where the two distributions seriously differ is Xhosa
(and, to a smaller degree, Turkish). This is caused
by the large number of word types that have occurrence 1 or 2
in the text. Therefore, it is likely that two consecutive words
have the same ranking and adding a small random number is now not negligible.
We ascribe this effect to the agglutinative nature of Xhosa.
Unlike, e.g., English, which expresses most of its syntactic functions
with isolated words, Xhosa displays agglutinated morphological
complexes. It is thus natural to expect in the Xhosa Bible
a large amount of hapax legomena. Any random shift will then represent
a significant perturbation to the original series, as observed in our data.

\section{List of literary works}\label{app_lit}

Table~\ref{tab_list} shows the full list of literary works
employed in the mesoscale and microscale analyses in Sec.~\ref{sec:results}.

\begin{table}
\begin{center}
\begin{ruledtabular}
\begin{tabular}{rll}
No. & Title & Author \\
\hline
1 & Andreas & Anonymous \\
2 & Anglo-Saxon Chronicle & Anonymous \\
3 & Beowulf & Anonymous \\
4 & Christ & Anonymous \\
5 & Genesis & Anonymous \\
6 & Guthlac & Anonymous \\
7 & Brut I & Layamon \\
8 & Brut II & Layamon \\
9 & Canterbury I & Chaucer \\
10 & Canterbury II & Chaucer \\
11 & Canterbury III & Chaucer \\
12 & Confessio Amantis I & Gower \\
13 & Confessio Amantis II & Gower \\
14 & Confessio Amantis III & Gower \\
15 & The Book of the Knight of La Tour-Landry & Caxton \\
16 & The Travels of Sir John Mandeville & Mandeville \\
17 & Every Man in His Humor. The Poetaster & Jonson \\
18 & Epicoene. Cynthia's Revels & Jonson \\
19 & Bartholomew Fair. The Alchemist & Jonson \\
20 & Tamburlaine the Great. Hero and Leander & Marlowe \\
21 & The Jew of Malta. The Massacre at Paris & Marlowe \\
22 & Paradise Lost & Milton \\
23 & Tragedies I & Shakespeare \\
24 & Tragedies II & Shakespeare \\
25 & Comedies I & Shakespeare \\
26 & Comedies II & Shakespeare \\
27 & The Innocence of Father Brown (1) & Chesterton \\
28 & The Man Who Knew Too Much (2) & Chesterton \\
29 & The Napoleon of Notting Hill (3) & Chesterton \\
30 & The Man Who Was Thursday (4) & Chesterton \\
31 & The Wisdom of Father Brown (5) & Chesterton \\
32 & Memoirs of Sherlock Holmes (6) & Conan Doyle \\
33 & The Return of Sherlock Holmes (7) & Conan Doyle \\
34 & The Sign of Four (8) & Conan Doyle \\
35 & The Hound of the Baskervilles (9) & Conan Doyle \\
36 & The Adventures of Sherlock Holmes (10) & Conan Doyle \\
37 & The Randolph Carter Stories (11) & Lovecraft \\
38 & The Dream Cycle (12)  & Lovecraft \\
39 & Twenty-Nine Tales (13) & Lovecraft \\
40 & Twenty-Nine Collaborative Stories  (14) & Lovecraft et al. \\
41 & At the Mountains of Madness (15) & Lovecraft \\
42 & The Works of Edgar Allan Poe I (16) & Poe \\
43 & The Works of Edgar Allan Poe II (17) & Poe \\
44 & The Works of Edgar Allan Poe III (18) & Poe \\
45 & The Works of Edgar Allan Poe IV (19) & Poe \\
46 & The Works of Edgar Allan Poe V (20) & Poe \\
\end{tabular}
\end{ruledtabular}
\end{center}
\caption{Literary works considered in the mesoscale and microscale analyses. The numerical identification of the left column is used in the legends of Figs.~\ref{fig_his} and the axes of Fig.~\ref{fig_his2} while the numbers in parentheses are employed for the axes of Fig.~\ref{fig_matrix}.}
\label{tab_list}
\end{table}

\section{Shuffled sentences} \label{app_shuf}
\begin{figure}[t]
\begin{center}
\includegraphics[width=0.49\textwidth, clip]{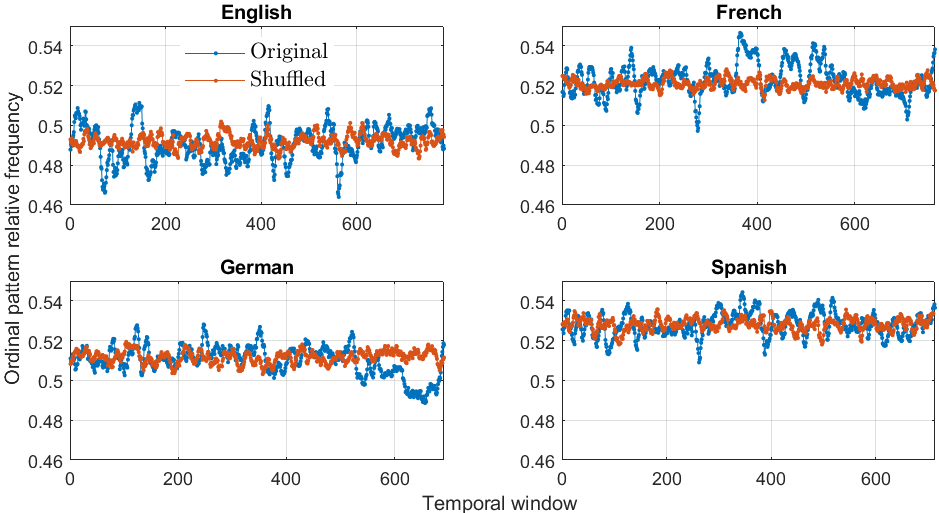}
\end{center}
\caption{Dynamical behavior for $D=2$ as in Fig.~\ref{fig_dyn} but comparing the original Bible (blue dots) and the Bible with shuffled sentences (red dots). In both cases, the curves differ from the case with shuffled words (the red curve in Fig.~\ref{fig_dyn}), which corresponds to the trivial dynamics.}
\label{fig_shuffdyn}
\end{figure}

\begin{figure}[t]
\begin{center}
\includegraphics[width=0.49\textwidth, clip]{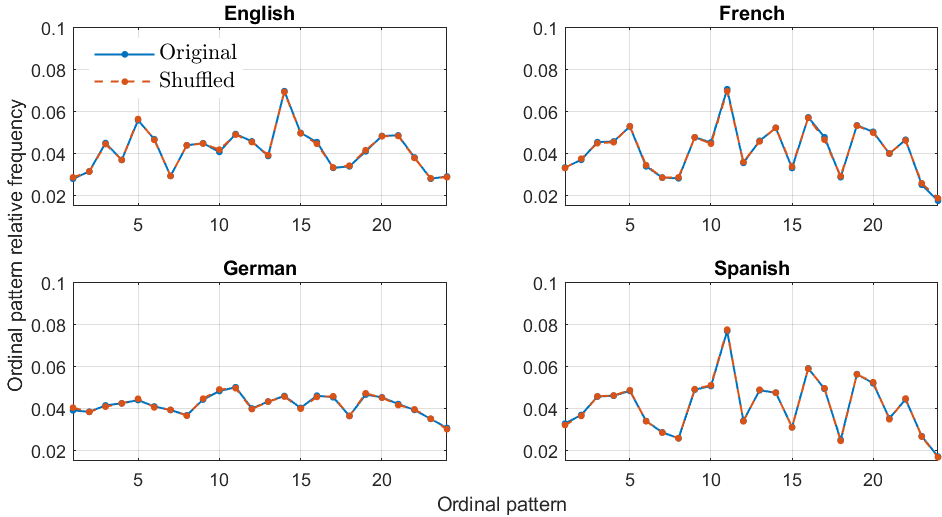}
\end{center}
 \caption{Pattern probability distributions for $D=4$ for the original Bible (blue lines as in Fig.~\ref{fig_bib}) and the Bible with shuffled sentences (red dots).}
\label{fig_shuffpat}
\end{figure}

The shuffled realizations of Figs.~\ref{fig_dyn}, \ref{fig_bib} and~\ref{fig_GC} 
are obtained by randomly shuffling all the words in the original
text. A different shuffled realization puts the sentences in random order,
instead of the individual words. Remarkably, our results obtained
for shuffled sentences are the same as those obtained for the original sequences.
For definiteness, we select four languages and plot in Fig.~\ref{fig_shuffdyn} both the original pattern dynamics for $D=2$ (blue dots),
reproduced from Fig.~\ref{fig_dyn}, and the ordinal
pattern when the sentences are shuffled (red dots).
Obviously, the dynamics do not agree because the relative frequencies
are calculated over time windows and these windows contain texts
with totally different sentences in both cases. However,
the stationary values and their probability distributions
are not 
modified. This is shown in Fig.~\ref{fig_shuffpat} for $D=4$,
where one can note that there is exact match between
the original Bible (blue lines as replicated
from Fig.~\ref{fig_bib}) and the Bible with shuffled sentences
(red dots). Since the short memory encountered in our analysis is based
on these statistical distributions, we can safely conclude that
our method detects short-range correlations
that typically occur inside a sentence.

\end{document}